\documentclass[letterpaper]{article} % DO NOT CHANGE THIS
\usepackage{aaai2026}  % DO NOT CHANGE THIS
\usepackage{times}     % DO NOT CHANGE THIS
\usepackage{helvet}    % DO NOT CHANGE THIS
\usepackage{courier}   % DO NOT CHANGE THIS
\usepackage[hyphens]{url}  % DO NOT CHANGE THIS
\usepackage{graphicx}      % DO NOT CHANGE THIS
\usepackage{natbib}        % DO NOT CHANGE THIS AND DO NOT ADD ANY OPTIONS TO IT
\usepackage{caption}       % DO NOT CHANGE THIS AND DO NOT ADD ANY OPTIONS TO IT

\usepackage{algorithm}
\usepackage{algorithmic}

\usepackage{newfloat}
\usepackage{listings}

\usepackage{amsmath}
\usepackage{amsfonts}
\usepackage{booktabs}
\usepackage{multirow}
\usepackage{pifont}
\usepackage{arydshln}
\DeclareCaptionStyle{ruled}{labelfont=normalfont,labelsep=colon,strut=off} % DO NOT CHANGE THIS
\floatstyle{ruled}
\newfloat{listing}{tb}{lst}{}
\floatname{listing}{Listing}
\title{\includegraphics[width=0.68cm]{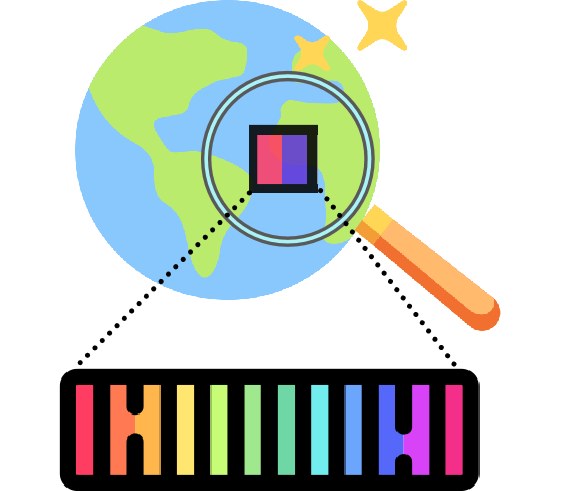} Earth-Adapter: Bridge the Geospatial Domain Gaps \\with a Frequency-Guided Mixture of Adapters}
\author{
    Xiaoxing Hu\textsuperscript{\rm 1}\equalcontrib,
    Ziyang Gong\textsuperscript{\rm 2}\equalcontrib,
    Yupei Wang\textsuperscript{\rm 1}\equalcontrib\thanks{Corresponding author},
    Yuru Jia\textsuperscript{\rm 3},
    Fei Lin\textsuperscript{\rm 4},\\
    Dexiang Gao\textsuperscript{\rm 5},
    Ke An\textsuperscript{\rm 1},
    Jianhong Han\textsuperscript{\rm 1},
    Zhuoran Sun\textsuperscript{\rm 1},
    Gen Luo\textsuperscript{\rm 6},
    Xue Yang\textsuperscript{\rm2$\dagger$}
}
\affiliations{
    \textsuperscript{\rm 1} Beijing Institute of Technology\\
    \textsuperscript{\rm 2} Shanghai Jiao Tong University\\
    \textsuperscript{\rm 3} KU Leuven\\
    \textsuperscript{\rm 4}  Macau University of Science and Technology\\
    \textsuperscript{\rm 5}  Peking University\\
    \textsuperscript{\rm 6}  Shanghai Artificial Intelligence Laboratory

}

\usepackage{bibentry}
\begin{document}

\maketitle

\begin{abstract}

Vision Foundation Models (VFMs), while powerful, often struggle in Remote Sensing (RS) segmentation tasks when combined with existing Parameter-Efficient Fine-Tuning (PEFT) methods. We observe that this limitation primarily arises from their inability to effectively handle the pervasive artifacts in RS imagery. To address this, we introduce Earth-Adapter, the first PEFT method specifically designed for RS artifact mitigation. Earth-Adapter introduces a novel Frequency-Guided Mixture of Adapters (MoA) approach, structured around a ``divide and conquer" strategy. It first utilizes Discrete Fourier Transformation (DFT) to "divide" features into distinct frequency components, thereby effectively isolating artifact-related information from semantic signals. Subsequently, to ``conquer" these artifacts, MoA independently optimizes features within different subspaces and dynamically assigns weights via a router to aggregate the refined representations. This enables adaptive refinement of the VFM’s representation space to mitigate the impact of artifacts. This simple yet highly effective PEFT method demonstrably mitigates artifacts and significantly enhances VFMs' performance on RS segmentation tasks. Extensive experiments demonstrate Earth-Adapter's effectiveness on in-domain semantic segmentation (SS), as well as Domain Adaptive (DA) and Domain Generalized (DG) semantic segmentation tasks. Compared with the baseline Rein, Earth-Adapter significantly improves mIoU by \textbf{1.2\%} in SS, \textbf{9.0\%} in DA, and \textbf{3.1\%} in DG benchmarks. Our code is available at https://github.com/VisionXLab/Earth-Adapter.
\end{abstract}

\section{Introduction}\label{sec:intro}

\begin{figure}[t!]
    \centering
    \includegraphics[width=0.9\linewidth]{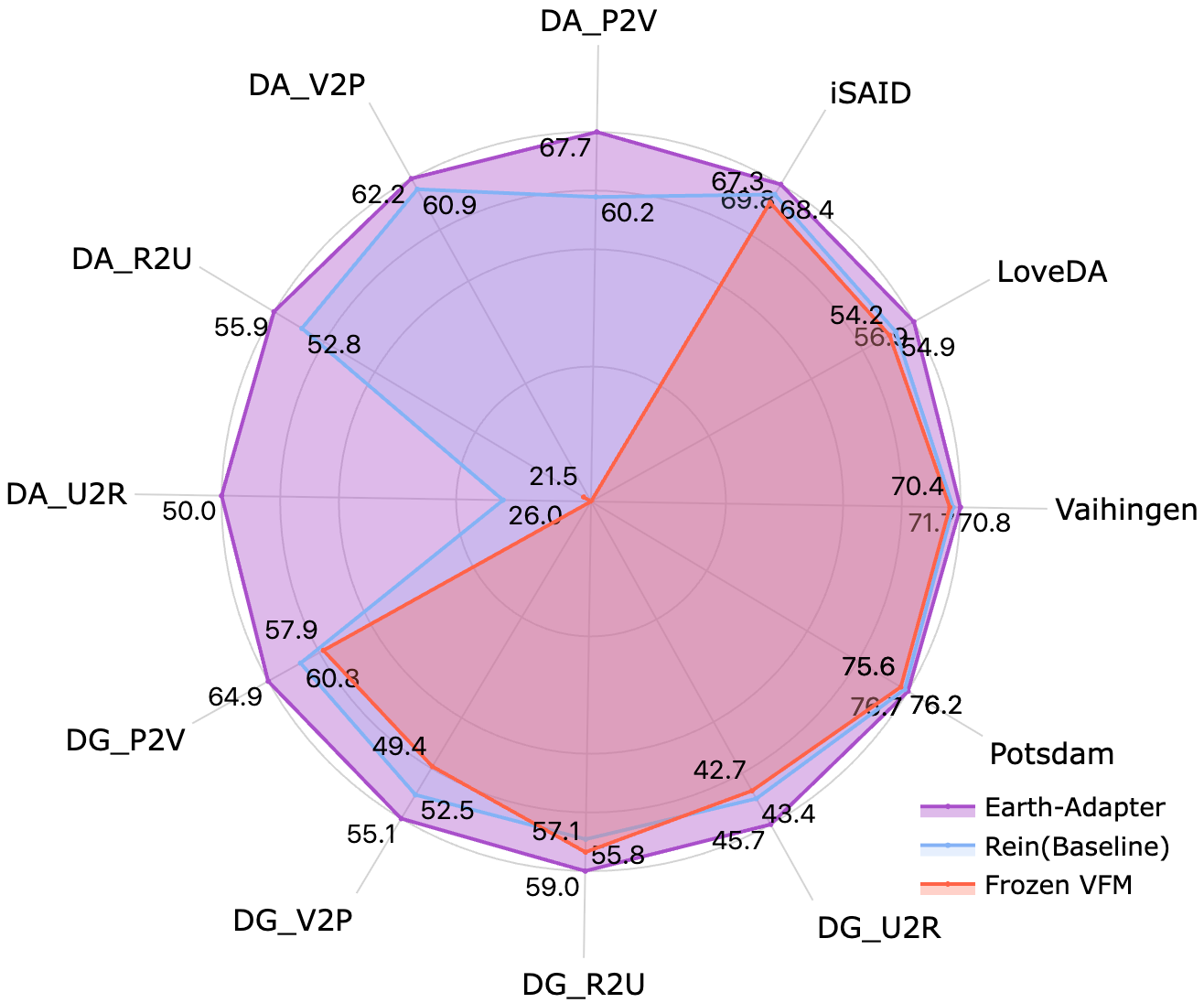}
    \caption{Performance across various remote sensing image segmentation benchmarks between Frozen VFM (DINOv2-L), Rein (Baseline) and the proposed Earth-Adapter.}
\end{figure}
Vision Foundation Models (VFMs) have made significant strides in recent years, driven by both natural language supervision paradigms (e.g., CLIP~\cite{clip}, ALIGN~\cite{align}, SigLIP~\cite{siglip}) and self-supervised paradigms (e.g., MAE~\cite{mae}, DINOv2~\cite{dinov2}). Pre-training on ultra-large-scale datasets empowers VFMs with robust zero-shot comprehension capabilities, leading to their extensive adoption across diverse tasks. These include classical computer vision tasks (classification~\cite{zhou2022learning}, detection~\cite{OV_det1,OV_det2}, segmentation~\cite{rsprompter}) as well as multimodal learning~\cite{llava}. Simultaneously, the conventional pre-train-then-fine-tune paradigm is evolving. When adapting VFMs to downstream tasks, the key challenge becomes how to efficiently preserve and unleash their inherent capabilities. Under this context, Parameter-Efficient Fine-Tuning (PEFT) methods \cite{lora, vpt, vanilla_adapter, coda} emerge as the pivotal solution due to their superior parameter-performance trade-off.

\begin{figure*}[t]
\centering
\includegraphics[width=1\linewidth]{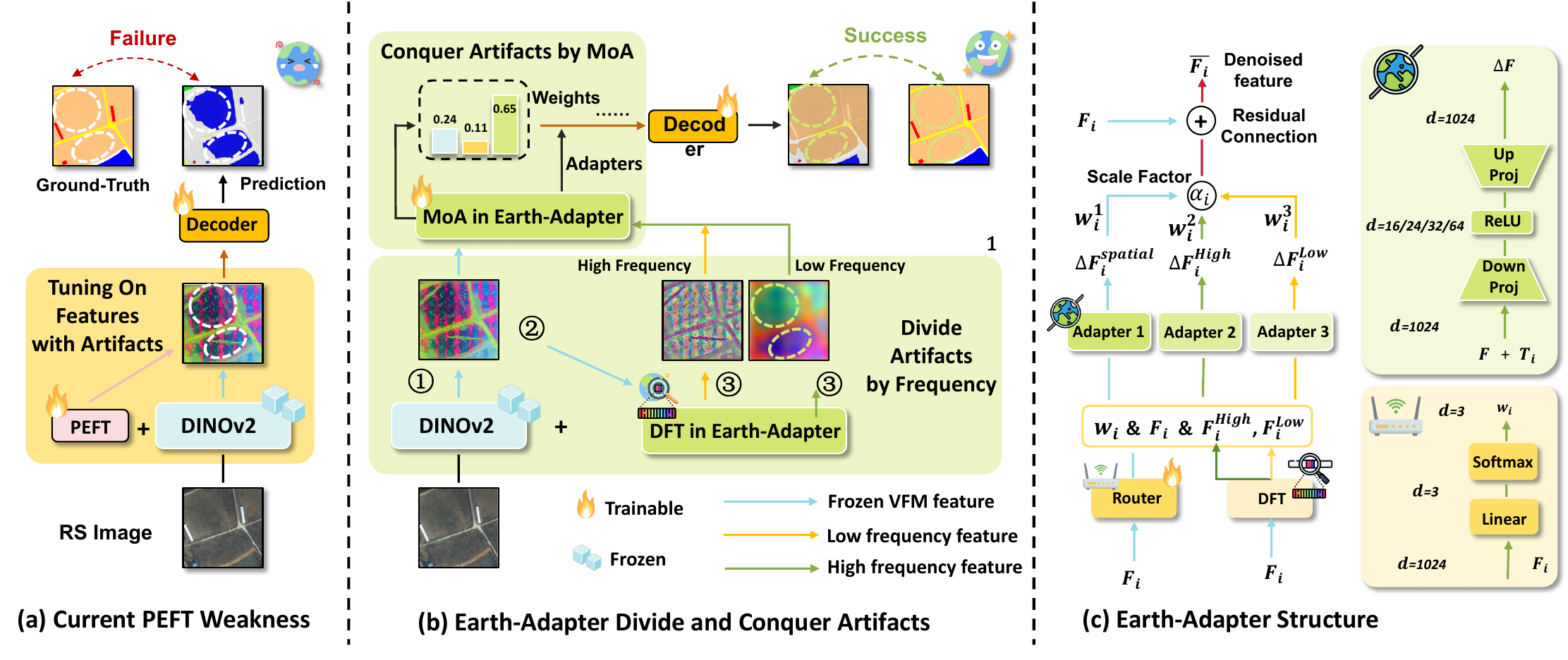}
\label{teaser}
\vspace{-0.5cm}
\caption{\textbf{Motivation and Structure Details of Earth-Adapter} (a) points out the artifact problems in existing PEFT methods. (b) illustrates how Earth-Adapter divides and conquers the artifacts by frequency-guided strategy and MoA framework. \textcircled{\scriptsize{1}}, \textcircled{\scriptsize{2}}, and \textcircled{\scriptsize{3}} show the sequence of each step in the DFT operation. (c) introduces the details of the Earth-Adapter component structures. }
\label{fig: motivation_and_structure}
\vspace{-8pt}
\end{figure*}

Existing PEFT methods designed for language or nature imagery have made great progress with excellent works, such as LoRA \cite{lora}, Visual Prompt Tuning (VPT) \cite{vpt}. However, most exhibit significant performance degradation in remote sensing (RS) segmentation tasks when integrated with DINOv2~\cite{dinov2} (experimental evidence provided in later sections). We argue the \textbf{\textit{main reason lies in the influences of artifacts in the features}}. As shown in Figure \ref{fig: motivation_and_structure} (a), we show the PCA visualization of features of DINOv2-L, which contains obvious redundant artifacts. We have also observed that artifacts in natural images exhibit distinct characteristics compared to those in RS images. In natural images, artifacts typically surround foreground objects \cite{vit_register}, such as humans or animals, and the disturbances they cause are relatively limited. In contrast, RS images, due to their overhead perspective, lack centralized subjects and contain multiple coexisting multi-scale targets. For example, a single RS image may simultaneously include large-scale agricultural regions and fragmented road networks. As a result, artifacts are almost situated everywhere in RS images shown in Figure \ref{fig: motivation_and_structure} (a), causing severe interference to pixel-level feature extraction,  which is pivotal for segmentation tasks.

To address artifacts and improve VFM performance in RS segmentation tasks, we propose a \textbf{``divide-and-conquer''} strategy. In the \textbf{``divide''} stage, based on the observation that high-frequency (HF) signals capture local details while low-frequency (LF) signals encode global structures, we apply Discrete Fourier Transformation (DFT) to separate HF and LF components, isolating artifacts in the HF domain. In the \textbf{``conquer''} stage, we design the Mixture of Adapters (MoA), which adjusts features in their respective frequency domains and then uses a router to dynamically assign weights for aggregating the corrected features. These aggregated features are further combined with the original frozen VFM features through skip connections and dynamic scaling coefficients, preserving the strong feature extraction ability of the underlying model.

This unique pipeline forms Earth-Adapter, a PEFT method tailored for RS semantic segmentation. We comprehensively evaluate Earth-Adapter on 12 established benchmarks across three settings—in-domain (SS), domain adaptive (DA), and domain generalized (DG)—and demonstrate its effectiveness. Compared with existing PEFT methods, Earth-Adapter achieves state-of-the-art (SOTA) performance on SS, DA and DG benchmarks and outperforming our baseline Rein~\cite{rein} by \textbf{1.2\%} , \textbf{9.0\%}  and \textbf{3.1\%}~mIoU, respectively. We also conduct an in-depth analysis of Earth-Adapter's design and potential. The core contributions can be summarized as follows:
\begin{itemize}
\item We observe that VFMs are prone to artifact interference in the high-dimensional feature space of RS imagery, which hinders their adaptation to RS segmentation tasks.

\item To address this, we propose Earth-Adapter, the first PEFT approach tailored specifically  to mitigate artifact-related issues.

\item We develop a Frequency-Guided Mixture of Adapters (MoA) that effectively isolates artifacts within distinct frequency subspaces, coupled with a dynamic router that adaptively fuses adapter outputs to optimize pixel-level VFM features for RS imagery.

\item Extensive experiments across diverse RS segmentation benchmarks demonstrate the effectiveness of Earth-Adapter.
\end{itemize}

\section{Related Work}\label{sec:related work}

\subsection{RS Semantic Segmentation}
Semantic segmentation~\cite{fcn, fcn_for_seg,deeplab}  aims to classify every pixel in an image. RS segmentation is a
critical task in numerous real-world applications, including land cover mapping, urban planning,
and environmental monitoring~\cite{9686686}. Cross-domain task is crucial due to the diversity of data, and lots of DA methods have been applied to improve the generalization ability of segmentation models\cite{rui2020survey,tong2020land}. Compared with DA, the number of DG research \cite{iizuka2023frequency, liang2024single} in RS is much less than DA. Although these works \cite{ zhu2023unsupervised} make contributions, most of them only focus on specialized models rather than leveraging the power of VFMs, leading to limited cross-domain capabilities. 

\subsection{Vision Foundation Models}
Vision Foundation Models (VFMs) have made significant progress and have exerted considerable influence across various AI-related fields. Among them, vision-language foundation models—such as CLIP~\cite{clip}, ALIGN~\cite{align}, SigLIP~\cite{siglip}, and InternViT~\cite{internvl}—are trained on billions of image-text pairs to learn joint visual-textual representations. These models have been widely applied to a broad range of vision and multimodal tasks, including classification~\cite{zhou2022learning}, detection~\cite{OV_det1,OV_det2}, segmentation~\cite{rsprompter}, image-text retrieval~\cite{clip}, image captioning~\cite{instructblip}, and large multimodal models (LMMs)~\cite{mono}. On the other hand, self-supervised models can also learn rich visual representations from large-scale unlabeled data. MAE~\cite{mae} learns generalized visual features by reconstructing masked images from partial inputs. MoCo~\cite{mocov3} employs a momentum encoder and contrastive learning to extract discriminative representations. DINOv2~\cite{dinov2} combines self-distillation with contrastive learning to produce scalable, transferable features for various downstream tasks. \textit{Beyond these various VFMs, a key challenge remains in effectively unlocking their potential to enhance performance on downstream tasks. Earth-Adapter is a pioneering work for RS segmentation tasks}

\subsection{PEFT in RS}
PEFT approaches, such as adapter \cite{vanilla_adapter, peft_nlp, lora} are proposed to fine-tune large pre-trained models to downstream tasks in Natural Language Processing (NLP) by introducing light-weight learnable parameters. Subsequently, the PEFT has begun to be transferred to vision\cite{yin2023vs, agiza2024mtlora} tasks and rapidly become the focus of researchers. Visual Prompt Tuning (VPT) \cite{vpt} first introduced `prompt tuning' into the vision area as learnable vectors. The thought of this paradigm is also similar to the adversarial re-programming \cite{adversarial_reprogramming}.
Currently, cross-domain works also focus on PEFT, such as \cite{first_prompt_learning_in_da} which is the first prompt tuning method in Domain Adaptation (DA) \cite{vpa, gao2022visual, coda} and \cite{rein} is the first work leveraging PEFT to fine-tuning VFMs for Domain Generalization (DG) \cite{bilearning}. In the RS field, the exploration of PEFT is also vigorous. For example, UPetu \cite{upetu} addresses storage for dense prediction tasks. TEA \cite{hu2024tea} uses an adapter network and top-down guidance for detection and segmentation tasks. \cite{lora_in_rs} also leverage LoRA to solve oriented detection task. However, for the RS semantic segmentation tasks, the research of PEFT is relatively blank, and existing methods can not handle the influence of artifacts. \textit{Earth-Adapter is the first PEFT designed to tackle artifacts, thereby unlocking the potential of VFMs for remote sensing segmentation tasks}.

\section{Method}
\subsection{Preliminary} 

We first denote the input RS image as \(\mathbf{x} \in \mathbb{R}^{H \times W \times 3}\), the corresponding label of \(\mathbf{x}\) as \(\mathbf{y} \in \mathbb{R}^{H \times W \times K}\), and the semantic segmentation prediction of \(\mathbf{x}\) as \(\mathbf{y}^\prime \in \mathbb{R}^{H \times W \times K}\), where \(K\) represents the number of semantic classes.

Here, we'll illustrate our innovative Earth-Adapter overview.

The proposed Earth-Adapter is composed of two key components: the MoA and Router. MoA consists of a \textit{Spatial Adapter}, a low-frequency (\textit{LF})\textit{ Adapter}, a high-frequency (\textit{HF}) \textit{Adapter}. We employ $R_\xi$, parameterized by $\xi$, to denote the router and $E^i_\epsilon, \; i \in \{1,2,3\}$, parameterized by $\epsilon$, to represent the three adapter experts. For the network architecture, we adopt the framework introduced by Rein~\cite{rein}, integrating DINOv2-L~\cite{dinov2} as the backbone, represented as $f_\phi$ and parameterized by $\phi$, alongside Mask2Former~\cite{mask2former} as the decoder, denoted by $f_\theta$ and parameterized by $\theta$. The following presents the details of Earth-Adapter, and we will elaborate on our training framework in the Training Framework section.

\subsection{Details of Earth-Adapter}\label{sec:earth-adapter}

Earth-Adapter aims to adapt VFMs to RS image semantic segmentation tasks with minimal learnable parameters.

\textbf{Optimization objective.} Before employing the Earth-Adapter, the optimization objective of fine-tuning VFMs is to identify a set of parameters that minimize the loss of the entire model on the downstream task:
\begin{equation}
    \mathop{\arg \min}_{\theta} \, 
    \sum_{(\mathbf{x}, \mathbf{y}) \in \mathcal{D}} 
    \mathcal{L}_{\text{seg}}\left( f_{\theta}(f_{\phi}(\mathbf{x})), \mathbf{y} \right),
\end{equation}
where $\mathcal{D}$ is the dataset and $\mathcal{L}_{seg}$ is the loss function, here we use the default CE loss. After incorporating our Earth-Adapter, the optimization objective becomes:
\begin{equation}\label{loss_fun1}
    \mathop{\arg \min}_{\theta, \epsilon, \xi}  
    \sum_{(\mathbf{x}, \mathbf{y}) \in \mathcal{D}} 
    \mathcal{L}_{\text{seg}}\left(f_{\theta}\left(
        \underbrace{R_\xi \circ E_\epsilon}_{\text{Earth-Adapter}}
        (f_{\phi}^*(\mathbf{x}))
    \right), \mathbf{y}\right).
\end{equation}
During training, the original parameters of the backbone (VFM) are kept frozen (denoted as $f_\phi^*$). In the following description, we denote the segmentation network combined with Earth-Adapter as $G_{\theta, \xi, \epsilon} $.

\textbf{Mixture of Adapters.}
Let $\mathbf{F}_i \in \mathbb{R}^{(hw) \times c}$ denote the visual feature at the $i$-th layer of the backbone, where $hw$ represents the token sequence length and $c$ denotes the token dimension. Our frequency adaptation operates through parallel processing streams:
The first \textit{Spatial  Adapter} employs a low-rank projection to refine the spatial feature:
\begin{equation}
\Delta\mathbf{F}^{spatial}_i = \text{Adapter}_1^i(\mathbf{F}_i^T),
\end{equation}
where $\Delta\mathbf{F}^{spatial}_i$ means the processed i-th layer Spatial-Domain features, and  $\textbf{Adapter}_1$ represents a nonlinear mapping layer composed of two low-rank matrices and an activation:
\begin{equation}\label{eq:adapter}
    \textbf{Adapter}_i = \textbf{W}_{up}(\text{Relu}(\textbf{W}_{down}(\cdot))).
\end{equation}
The \textit{Frequency Adapters} consist of a \textit{HF Adapter} (high-frequency subspace) and a \textit{LF Adapter} (low-frequency subspace), which fine-tune features in specific frequency subspaces derived from a 2D Discrete Fourier Transform (DFT) decomposition. We first reshape $\mathbf{F}^{spatial}$ to $({C, H, W})$ and apply DFT on spatial features as $\mathcal{FT} (\mathbf{F}_{spatial})$. When splitting the frequency domains, we employ a fixed frequency cutoff $\rho$ to decompose the spectrum into high and low frequency components. Subsequently, these components are transformed back into features via the Inverse Fourier Transform (IFT), yielding \textit{LF} and \textit{HF} features:

\begin{equation}
\mathbf{F}^{low}_i = \mathcal{FT}^{-1}(\mathbf{M} \odot \mathcal{FT}(\mathbf{F}_{spatial})),
\end{equation}
\begin{equation}
\mathbf{F}^{high}_i = \mathcal{FT}^{-1}((1-\mathbf{M}) \odot \mathcal{FT}(\mathbf{F}_{spatial})).
\end{equation}
The frequency mask $\mathbf{M} \in \{0,1\}^{H \times W}$ is defined by:
\begin{equation}
\mathbf{M}(u,v) = \begin{cases} 
1 &, \thinspace\text{if } \max(|u-\frac{H}{2}|, |v-\frac{W}{2}|) \leq \rho\frac{H}{2} \\
0 &, \thinspace\text{otherwise} 
\end{cases}
\end{equation}
Thereafter, the \textit{LF} and \textit{HF} features are independently passed through two distinct low-rank linear projection layers, generating frequency adaptation adjustments:
\begin{equation}
\Delta\mathbf{F}^{low}_i = \text{Adapter}_2^i(\mathbf{F}^{low}_i),
\end{equation}
\begin{equation}
\Delta\mathbf{F}^{high}_i = \text{Adapter}_3^i(\mathbf{F}^{high}_i).
\end{equation}
The \textit{Frequency Adapters} shares the same structure as the \textit{Spatial Adapter}, as defined in Equation~\ref{eq:adapter}.

\textbf{Dynamic Router.} We implement dynamic feature aggregation through a Router that learns optimal combinations of feature adjustment according to the original visual features. The Router weight is computed by channel-wise attention:
\begin{equation}
\mathbf{w}_i = \text{Softmax}(R_\xi(\mathbf{F}_i)),
\end{equation}
where $\mathbf{w}_i$ represents weights for \textit{spatial}, \textit{LF}, and \textit{HF} components.
The final feature adjustment is calculated as:
\begin{equation}
\Delta\mathbf{F}_i = \alpha_i \sum_{k=1}^3 \mathbf{w}_i^{(k)} \Delta\mathbf{F}^{(k)}_i,
\end{equation}
where $\alpha_i$ is a learnable scaling parameter with small initial value, and $k \in \{spatial, low, high\}$.
The frozen features and the refined features are fused via a skip connection:

\begin{equation}
\bar{\mathbf{F}_{i}} = \mathbf{F}_i + \Delta\mathbf{F}_i.
\end{equation}
$\bar{\mathbf{F}_i}$ is then forwarded to the subsequent Transformer block to continue the layer-wise processing.
\subsection{Training Framework}\label{sec:training framewotk}
In this work, we explore the application of Earth-Adapter across three distinct semantic segmentation subtasks: semantic segmentation (SS), domain adaptive (DA) semantic segmentation and domain generalized (DG) semantic segmentation .
For SS and DGSS, we train Earth-Adapter with Mask2former~\cite{mask2former} in end-to-end supervised learning as described in Equation~\ref{loss_fun1}.
For DA semantic segmentation, we use the self-training framework introduced in DACS~\cite{dacs} without using target domain labels. DACS employs an EMA teacher to generate pseudo labels for target-domain images, mixes target and source samples to obtain the mixed domain $(\mathbf{x}_{mix}, \mathbf{y}_{mix})$ and then jointly trains on both the source and mixed domains.
The optimization objective for the DASS if defined by:
\begin{equation}
    \begin{aligned}
        \mathop{\arg \min}_{\theta, \epsilon, \xi} \, & \sum_{(\mathbf{x}, \mathbf{y}) \in \mathcal{D} \cup \mathcal{D}_{mix}} \mathcal{L}_{\text{seg}}\left(G_{\theta, \xi, \epsilon}(\mathbf{x}), \mathbf{y}\right) \\
        & \quad + \lambda_{da} \mathcal{L}_{\text{seg}}\left(G_{\theta, \xi, \epsilon}(\mathbf{x}_{mix}), \mathbf{y}_{mix}\right).
    \end{aligned}
\end{equation}
$\lambda_{da}$ is the weight parameter for DA loss.
\section{Experiments}\label{sec:exp}

\begin{table*}[h]
\centering
\setlength{\tabcolsep}{8pt} % 设置列间距
% \fontsize{7pt}{7pt}\selectfont
\begin{tabular}{lcccccccccc}
\toprule
\multirow{2}{*}{\textbf{Methods}} & \multicolumn{4}{c}{\textbf{DASS}} & \multirow{2}{*}{\textbf{Avg.}} & \multicolumn{4}{c}{\textbf{DGSS}} & \multirow{2}{*}{\textbf{Avg.}} \\
\cmidrule(lr){2-5} \cmidrule(lr){7-10}
& \textbf{P2V} & \textbf{V2P} & \textbf{R2U} & \textbf{U2R} & & \textbf{P2V} & \textbf{V2P} & \textbf{R2U} & \textbf{U2R} & \\
\midrule
\textit{Previous SOTA methods} \\
\midrule
DAFormer  & 64.4 & 54.8 & 52.7 & 42.5 & 53.6 & 42.4 & 41.4 & 54.2 & 39.9 & 44.5 \\
HRDA& \underline{67.6} & 58.6 & 53.2 & 35.3 & 53.7 & 33.1 & 31.1 & 54.2 & 39.8 & 39.6 \\
\midrule
\textit{PEFT-based methods with VFM} \\
\midrule
Frozen & 21.0 & 7.2 & 21.5 & 11.8 & 15.4 & 57.9 & 49.4 & 57.1 & 42.7 & 51.8 \\
Frozen (with register) & 11.3 & 28.2 & 36.4 & 16.1 & 23.0 & 59.0 & 51.1 & 58.9 & 44.9 & 53.5 \\
Full Fine-Tune & 11.3 & 16.5 & 23.1 & 10.6 & 15.4 & 12.6 & 19.6 & 33.1 & 19.7 & 21.3 \\
Adapter  & 66.4 & 59.3 & \textbf{55.9} & 46.2 & 57.0 & 56.0 & 47.4 & \underline{57.6} & 41.8 & 50.7 \\
LoRA  & 17.8 & 18.8 & 24.5 & 26.0 & 15.7 & 20.3 & 25.6 & 29.3 & 21.9 & 24.3 \\
VPT  & 66.2 & 59.3 & \underline{55.3} & \underline{48.0} & \underline{57.2} & 59.2 & 52.3 & 57.4 & \underline{44.9} & \underline{53.5} \\
AdaptFormer  & 12.9 & 15.3 & 25.1 & 22.2 & 18.9 & 14.0 & 19.3 & 31.0 & 15.8 & 20.0 \\
Rein (baseline)  & 60.2 & \underline{60.9} & 52.8 & 26.0 & 50.0 & \underline{60.8} & \underline{52.5} & 55.8 & 43.4 & 53.1 \\
\textbf{Earth-Adapter (Ours)} & \textbf{67.7 } & \textbf{62.2 } & \textbf{55.9 } & \textbf{50.0 } & \textbf{59.0 } & \textbf{64.9 } & \textbf{55.1 } & \textbf{59.0} & \textbf{45.7 } & \textbf{56.2 } \\
\midrule
\textit{$\Delta$ over baseline method} & \textbf{+7.5} & \textbf{+1.3} &\textbf{+3.1} & \textbf{+24.0} & \textbf{+9.0} & \textbf{+4.1} & \textbf{+2.6} & \textbf{+3.2} & \textbf{+2.3} & \textbf{+3.1} \\
\bottomrule
\end{tabular}
\caption{\textbf{Performance (mIoU\%) comparison between previous SOTA methods and PEFT-based methods on DA and DG benchmarks.} \textbf{Bold} indicates the best performance. Underlined results denote the second-best. The last row shows improvements over the baseline.}
\end{table*}

\begin{table}[t!]
    \centering
    \setlength{\tabcolsep}{2mm} % 设置列间距
    \fontsize{9pt}{9pt}\selectfont
    \begin{tabular}{lccccc}
    \toprule
         \textbf{Method} & \textbf{P} & \textbf{V} & \textbf{L}& \textbf{i} & \textbf{Avg.} \\
    \midrule
    \multicolumn{6}{l}{\textit{Traditional Methods}} \\
    \midrule
    DeepLabV3+ (R-101) & 74.8 & 69.7 & 51.4 & 53.4 & 62.3 \\
    PSPNet (R-101)     & 74.7 & 69.2 & 48.5 & 57.5 & 62.5 \\
    UperNet (Swin-B)     & 75.5 & 69.4 & 55.6 & 67.9 & 67.1 \\
    Segformer (MiT-B5)      & 75.3 & 68.4 & 55.5 & 66.8 & 66.5 \\
    Mask2Former (Swin-B) & 75.7 & 

    \underline{71.2} & 54.7 & 64.0 & 66.4 \\
    \midrule
    \addlinespace
    \multicolumn{6}{l}{\textit{PEFT-based methods with VFM}} \\
    \midrule
    Frozen                  & 75.6 & 70.4 & 54.2 & 67.3 & 66.9 \\
    Frozen (with register) & 75.7 & 70.8 & \underline{56.0} & 67.5 & 67.5 \\
    Full Fine-Tune         & 70.5 & 64.6 & 43.9 & 12.0 & 47.7 \\
    LoRA                   & 71.9 & 64.6 & 51.0 & 48.5 & 59.0 \\
    VPT                    & 75.9 & 70.6 & 54.6 & \underline{68.5} & 67.4 \\
    AdaptFormer            & 65.4 & 56.0 & 42.0 & 7.5  & 42.8 \\
    Rein                   & \underline{76.2} & 70.8 & 54.9 & 68.4 & \underline{67.6} \\
    \textbf{Earth-Adapter (Ours)}          & \textbf{76.7} & \textbf{71.7} & \textbf{56.9} & \textbf{69.8} & \textbf{68.8} \\
    \addlinespace
    \midrule
    \textit{$\Delta$ over baseline method} & 
    \textbf{+0.5} & \textbf{+0.9} & \textbf{+2.0} & \textbf{+1.4} & \textbf{+1.2}\\
    \bottomrule
    \end{tabular}
    \caption{\textbf{Performance (mIoU\%) comparison between previous SOTA methods and PEFT-based methods on SS benchmarks.} \textbf{Bold} indicates the best performance. Underlined results denote the second-best. The last row shows improvements over the baseline.}
    \label{tab:SS_bench}
\end{table}

%isaid跑的比较慢，最后再跑
\begin{table}[h]
    \centering
    \setlength{\tabcolsep}{1.2mm} % 设置列间距
    \fontsize{8.5pt}{8.5pt}\selectfont
    \begin{tabular}{lccccc}
        \toprule
        \multicolumn{6}{l}{\textit{General VFMs}} \\
        \midrule
         Backbone  & \multirow{1}{*}{Methods} & Params & P2V(DG) & V2P(DG) & Avg.  \\
        \hdashline
          \multirow{2}{*}{\begin{tabular}[c]{@{}l@{}}DINOv2-S\end{tabular}}&Frozen & 0.0M  & 44.3 & 43.3 & 43.8 \\
         &  Earth-Adapter &  2.6M\textasciitilde9.6M &  \textbf{47.5} &  \textbf{44.8} &  \textbf{46.2}  \\
         \hdashline
          \multirow{2}{*}{\begin{tabular}[c]{@{}l@{}}DINOv2-B \end{tabular}}&Frozen & 0.0M & 51.4 & 49.5 & 50.5\\
         &  Earth-Adapter &  2.6M\textasciitilde9.6M &  \textbf{53.0} &  \textbf{ 51.8} & \textbf{52.4} \\
         \hdashline
         \multirow{2}{*}{\begin{tabular}[c]{@{}l@{}}DINOv2-L\end{tabular}}&Frozen & 0.0M & 57.9 & 49.4 & 53.7 \\
         &  Earth-Adapter &  2.6M\textasciitilde9.6M &  \textbf{64.9} &  \textbf{55.1}& \textbf{60.0} \\
         \midrule
        \multicolumn{6}{l}{\textit{RS-pretrained VFMs}} \\
        \midrule
                 Backbone  & \multirow{1}{*}{Methods} & Params & P2V(DA) & V2P(DA) & Avg.  \\
        \hdashline
          \multirow{2}{*}{\begin{tabular}[c]{@{}l@{}}MTP-L\end{tabular}}&Frozen & 0.0M  & 32.0 & 33.0 & 32.5 \\
         &  Earth-Adapter &  2.6M\textasciitilde9.6M &  \textbf{40.8} &  \textbf{41.0} &  \textbf{40.9}  \\
         \hdashline
          \multirow{2}{*}{\begin{tabular}[c]{@{}l@{}}ScaleMAE-L \end{tabular}}&Frozen & 0.0M & 20.0 & 19.8& 19.9\\
         &  Earth-Adapter &  2.6M\textasciitilde9.6M &  \textbf{20.2} &  \textbf{ 29.6} & \textbf{24.9} \\
         \hdashline
         \multirow{2}{*}{\begin{tabular}[c]{@{}l@{}}DOFA-L\end{tabular}}&Frozen & 0.0M & 9.9 & 13.7 & 11.8 \\
         &  Earth-Adapter &  2.6M\textasciitilde9.6M &  \textbf{26.2} &  \textbf{33.2}& \textbf{20.7} \\
        \bottomrule
    \end{tabular}
    \caption{\textbf{Ablation studies of Earth-Adapter on different VFMs.}  Results validate that Earth-Adapter's effectiveness on diverse backbone networks.}
    \label{tab: backbone_ablation}
\end{table}

\begin{figure*}[t]
\centering
\includegraphics[width=1\linewidth]{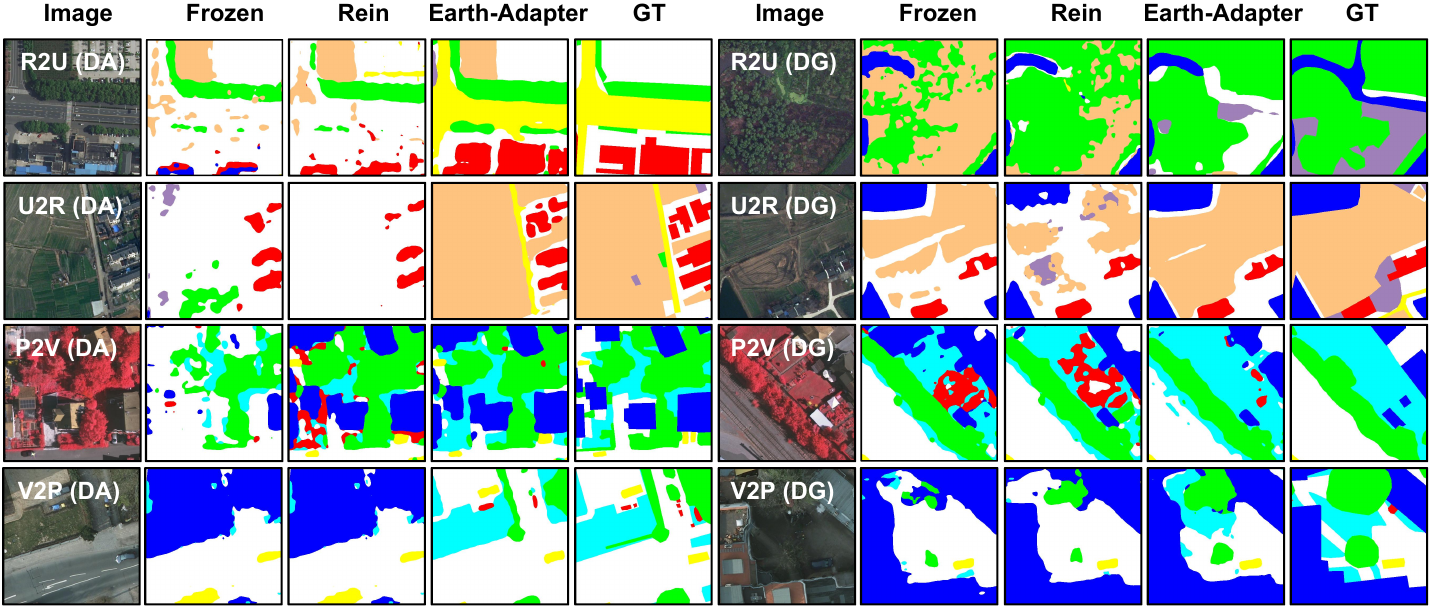}
\label{fig: seg_map}
\vspace{-0.5cm}
\caption{\textbf{Visualization of Predicted Segmentation Maps} We Compare Earth-Adapter with the Frozen DINOv2-L backbone and our baseline Rein on eight cross-domain benchmarks. For the Potsdam and Vaihingen color map, white is the Impervious surface,red is the clutter, blue is the building, Cyan is the low vegetation, green is the tree, and yellow is the car. For LoveDA color map,  red is the building, yellow is the road, blue is the water, purple is the barren, green is the forest, brown is the agriculture.}
\label{fig: seg map}
\end{figure*}

\subsection{Experimental setup}
\textbf{Datasets and Benchmarks.} We conduct all experiments on several widely used RS image segmentation datasets: Potsdam~\cite{isprs} , Vaihingen~\cite{isprs}, LoveDA~\cite{loveda}, and iSAID~\cite{isaid}.
Following the standard MMSegmentation~\cite{mmseg2020} configurations, we split each dataset into training and validation sets, apply cropping, and construct four semantic segmentation (SS) benchmarks: \textbf{Potsdam}(P), \textbf{Vaihingen}(V),\textbf{LoveDA}(L),\textbf{iSAID}(i), along with four domain adaptation (DA) and four domain generalization (DG) tasks: \textbf{Potsdam} to \textbf{Vaihingen} (P2V), \textbf{Vaihingen} to \textbf{Potsdam} (V2P), \textbf{Rural} to \textbf{Urban} (R2U), \textbf{Urban} to \textbf{Rural} (U2R).
In our experimental setup, all images are cropped to a size of $512\times512$.

\textbf{Implementation Details.}
We use MMSegmentation as our training and evaluation framework. We employ Mask2former~\cite{mask2former} as our decoder, which is a highly efficient and effective algorithm for semantic segmentation tasks. For backbone, we utilize the Dinov2-Large~\cite{dinov2} model to extract features, which are subsequently utilized as input for the Mask2former. During training, we utilize AdamW~\cite{adamw} as the optimizer, with a learning rate of 1e-5 for the backbone, 1e-4 for the decoder, and 1e-4 for the relevant parameters in PEFT. Within the DACS training framework, we set $\lambda_{da}$ to 0.5.

\subsection{Main Result}
% \textbf{Cross-Domain Performance}
As depicted in Table 1, we conduct a comparative analysis of the existing mainstream PEFT methods, such as Adapter~\cite{vanilla_adapter}, LoRA~\cite{lora}, and VPT~\cite{vpt}, against our Earth-Adapter across eight cross-domain benchmarks,with state-of-the-art method Rein~\cite{rein} in natural scene as our baseline. These benchmarks encompass four DA scenarios (P2V, V2P, R2U, and U2R) and four DG scenarios that mirror the DA ones.
\begin{figure*}[t]
\centering
\includegraphics[width=1\linewidth]{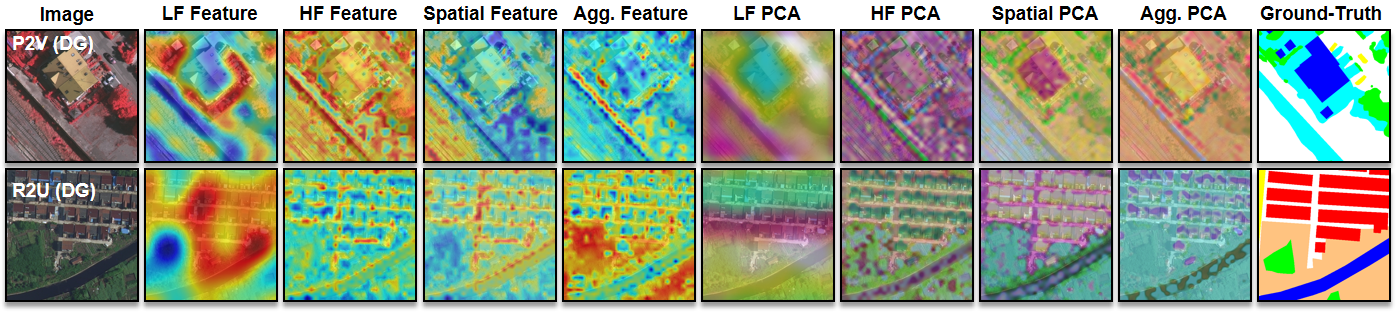}
\vspace{-0.5cm}
\caption{\textbf{Visualization and PCA of Adapters' Feature Maps.} `Agg. Feature' represents the aggregated adapters' features. `PCA' represents the Principal Component Analysis of features. All visualizations represent feature maps, not heatmaps. Thus only the semantic boundaries within the features should be focused rather than color intensities.}
% \label{fig: motivation_and_structure}
\label{fig: pca}
\end{figure*}
%  \begin{figure}[t]
% \centering
% \includegraphics[width=1\linewidth]{images/cut_off.png}
% % \label{fig: dimension_cut_off}
% \vspace{-0.6cm}
% \caption{\textbf{Parameter-sensitive experiments of the cut-off frequency of DFT.} The results on P2V (DA) and U2R (DG) benchmarks show Earth-Adapter's robustness to the cut-off frequency.}
% \label{fig: cut_off}
% \end{figure}

Starting with the DA experiments, many existing PEFT methods suffer from severe performance degradation when compared to DA-specialized models. For example, LoRA achieves only 17.8\% mIoU on the P2V (DA) benchmark for RS images, even lower than the Frozen DINOv2-L backbone. In contrast, our Earth-Adapter demonstrates much stronger domain adaptation capabilities. It surpasses the baseline Rein by 7.5\% mIoU on P2V, 3.1\% mIoU on R2U, 24.0\% mIoU on U2R, and achieves an average improvement of 9.0\% mIoU across all DA benchmarks. Shifting to the DG experiments, the issue of performance degradation remains, although the trend is somewhat mitigated. Unlike in the DA setting, the Frozen model highlights the potential generalizability of DINOv2-L. However, FFT, LoRA, and AdaptFormer continue to show clear limitations. Consistent with the DA benchmarks, Earth-Adapter achieves state-of-the-art (SOTA) performance in DG, surpassing the baseline Rein and Adapter by 3.1\% and 4.4\% mIoU, respectively.

We also conducted comparisons on four SS benchmarks, with the results summarized in Table 2. Regarding PEFT-related methods, conventional approaches such as LoRA and AdaptFormer struggle to adapt to the characteristics of remote sensing images, often disrupting the representations of VFMs during fine-tuning and ultimately leading to poor performance. In contrast, our Earth-Adapter consistently achieves the best performance among all PEFT methods. We attribute this advantage to its unique design, which mitigates high-dimensional artifacts and effectively "denoises" semantic representations, thereby enabling more accurate pixel-level feature extraction and superior semantic segmentation performance. Overall, our Earth-Adapter surpasses the baseline Rein by an average of 1.2\% mIoU, highlighting its effectiveness.

\subsection{Ablation and Analysis}

\begin{table*}[h]
    \centering
    
    \begin{tabular}{ccccccc}
    \toprule
    \textbf{Backbone}& \multirow{1}{*}{\textbf{Methods}} & \textbf{Params} $\downarrow$ & \textbf{Train Speed (s/iter)} $\downarrow$ & \textbf{Infer Speed (s/iter)} $\downarrow$ & \textbf{mIoU (\%) $\uparrow$}\\
        \midrule
          \multirow{3}{*}{DINOv2-L}& Full & 304.2M & 0.80 & 0.11   & 15.4\\
          &Rein & 3.0M & 0.68 & 0.12    & 50.0\\
          &Earth-Adapter &  2.6M\textasciitilde9.6M &  0.71 & 0.12  & \textbf{59.0 (+9.0)} \\
        \bottomrule
    \end{tabular}

    \caption{\textbf{Comparison on Parameters, Speed, and Performance between FFT, Rein, and Earth-Adapter.} All results are the average score of DA benchmarks, demonstrating the excellent trade-off of Earth-Adapter.}
    \label{tab: trade-off}
\end{table*}
\begin{table}[h]
    \centering
        \setlength{\tabcolsep}{1.2
        mm} % 
    \fontsize{9pt}{9pt}\selectfont
    \begin{tabular}{cccccc}
        \toprule
          \textbf{Adapter} & + \textbf{ \textit{HF}} & + \textbf{ \textit{LF}}  & \textbf{P2V (DG)}& \textbf{V2P (DG)}& \textbf{Avg.}  \\
         \midrule
          1 & - & - & 61.0 & 52.8 & 56.9 (+0.0)\\
          - & 1 & - &  59.4 & 50.9 &50.9 (-6.0)  \\
         - & - & 1 & 59.1 & 51.5 & 55.3 (-1.6) \\
        \midrule
         2 & - & - & 61.6  & 51.4  & 56.5 (-0.4) \\%
          3 & - & - & 64.0 & 52.7 & 58.4 (+1.5)\\
          4 & - & - & 61.4 &  53.3 & 57.4 (+0.5) \\
         5 & - & - &62.5  & 52.5 & 57.5 (+0.6) \\
        \midrule
         1 & 1 & 1 & \textbf{64.9} &  \textbf{55.1}&  \textbf{60.0} (+3.1)\\
        2 & 1 & 1 & 61.3 & 54.7 & 58.0 (+1.1)\\
         3 & 1 & 1 & 61.6  & 54.6 & 58.1 (+1.2)\\
        
        % 1 & 1&1 &1 & 1 &1\\
        \bottomrule
    \end{tabular}
    \caption{\textbf{Ablation of Adapters' Combination.} The `Adapter' means the number of vanilla Adapter that accepts \textit{Spatial} features. The `+ \textit{HF}' and `+ \textit{LF}' represent the number of adapters that accept high and low-frequency features as input. The `+' in the bracket represents the gain compared with one \textbf{Adapter}.}
    \label{tab: component ablation}
\end{table}
% \begin{figure}[t]
% \centering
% \includegraphics[width=1\linewidth]{images/dimension.png}
% \vspace{-0.5cm}
% \caption{\textbf{Parameter-sensitive experiments of the projector dimension in Adapter.} The results on P2V (DA) and U2R (DG) benchmarks show Earth-Adapter's robustness to dimensions.}
% \label{fig: dimension}
% \end{figure}

\textbf{Ablation of Different Backbones.} Beyond the DINOv2-Large backbone used in the main results (Tables 1 and 2), we further evaluated other DINOv2 variants. Results for ViT-Small and ViT-Base (Table 3, rows 3–8) show that Earth-Adapter consistently improves performance across backbones of different scales. We also tested VFMs pre-trained on remote sensing data, including MTP-Large~\cite{mtp}, ScaleMAE-Large~\cite{scale_mae}, and DOFA-Large~\cite{dofa} (Table 3, rows 11–16). Earth-Adapter again delivers stable improvements in segmentation performance. Finally, we observed that DINOv2 models outperform those pre-trained on remote sensing data, likely because the latter are trained on smaller-scale datasets and thus remain less mature.

\textbf{Component Effectiveness Ablation.}
In Table 4, ablation studies on MoA's components and adapter numbers show that using only \textit{HF} or \textit{LF} adapters decreases performance compared to spatial adapters, with \textit{HF} causing a larger drop. This aligns with PCA visualizations in \ref{fig: motivation_and_structure} (b) and \ref{fig: pca}, where \textit{HF} features have more artifacts and noise, while \textit{LF} features are smooth with clear global semantics. Thus, using only \textit{HF} severely damages performance. Although only using \textit{LF} features outperforms \textit{HF} features (50.9\%mIoU), it still underperforms \textit{spatial} features (56.9\% mIoU) due to lacking semantic details. After that, we also conduct more experiments on the number of \textit{Spatial} adapters, and reveal the best composition is the one \textit{spatial} adapter with one \textit{HF} and \textit{LF} adapters.

\textbf{Prediction Visualization Analysis.}
In Figure \ref{fig: seg map}, we visualize the predicted segmentation map of the Frozen DINOv2-L, Rein, and Earth-Adapter.  It is worth noting that in the U2R (DG) example, Rein's prediction is worse than the original backbone, which means Rein can not adapt to RS image well, which leads to a negative effect on the backbone features. In contrast, on the backbone's basement, the Earth-Adapter keeps the good details of backbone features, and further optimizes the representation of agriculture class. And in other experiments, Earth-Adapter also exhibits better prediction than Rein and Frozen backbone, showing a higher performance upperbound.

\begin{table}[t!]
\centering
\begin{tabular}{lccc}
\toprule
\textbf{Method} & \textbf{P2V(DG)} & \textbf{V2P(DG)} & \textbf{Avg.} \\
\hline
w/o DR & 63.6 & 53.2 & 58.4 \\
w/ DR  & \textbf{64.9} & \textbf{55.1} & \textbf{60.0(+1.6)} \\
\bottomrule
\end{tabular}
\caption{\textbf{Ablation of Dynamic Router.}}
\label{tab:dr}
\end{table}

\textbf{Feature Visualization Analysis.} In Figure \ref{fig: pca}, we visualize features captured by three adapters on P2V (DG) and R2U (DG) benchmarks. As discussed in the introduction, \textit{LF} features focus on coarse-grained and global semantics, while \textit{HF} features show detailed representations. The aggregated features are a weighted summation of the three different frequency features. PCA visualizations clearly show each feature's characteristics. Take P2V in the figure as an example, artifacts are almost filtered into high-frequency features (shown in `\textit{HF} PCA'), and dynamic fusion of all features ensures the final aggregated features' PCA maintains clear semantic edges and successfully filters out artifacts. These visualizations further enhance Earth-Adapters' explainability.

\textbf{Performance-Speed Trade-Off Analysis.} We compare the parameters, speed, and performance between FFT, Rein, and Earth-Adapter in Table \ref{tab: trade-off}. Significantly, at the same time of close training and inference speed with Rein, Earth-Adapter achieves better performance (9.0\% mIoU improvement), with a smaller parameter scale. This further confirms the efficiency of Earth-Adapter, which can be attributed to its simple yet effective design methodology, making it particularly well-suited for semantic segmentation tasks on remote sensing images.

\textbf{Ablation of Dynamic Router.} As shown in Tab.~\ref{tab:dr}, the use of the dynamic router leads to better results, yielding an average improvement of 1.6\% mIoU across the two DG benchmarks. Compared with static weights (where we assign a weight of 1/3 to each static route), the dynamic router can adaptively adjust the feature allocation weights, enabling more effective representation optimization.

\section{Conclusion}

Vision Foundation Models (VFMs) excel across diverse visual tasks, and pairing them with Parameter-Efficient Fine-Tuning (PEFT) is an effective way to adapt them to downstream applications. However, existing PEFT methods often fall short on remote sensing (RS) semantic segmentation because they fail to suppress the pervasive artifacts in VFM deep features for RS imagery. To address this limitation, we introduce Earth-Adapter, a simple yet effective PEFT framework tailored for RS segmentation. It employs a frequency-guided mixture of adapters (MoA) that isolates artifacts into the high-frequency subspace, fine-tunes features within each subspace, and adaptively fuses them through a learnable router.

% Uncomment the following to link to your code, datasets, an extended version or similar.
% You must keep this block between (not within) the abstract and the main body of the paper.
\clearpage

\section{Acknowledgments}
This work was partly supported by National Natural Science Foundation of China (62301046, 62506229) and Natural Science Foundation of Shanghai (25ZR1402268).

\section{Appendix}
\section{Training Details}
\subsection{Dataset and Benchmark}
\begin{table*}[t]
    \centering
    % \setlength{\tabcolsep}{2mm} % 设置列间距
    % \fontsize{9pt}{9pt}\selectfont
    \begin{tabular}{ccccc}
    \toprule
    Dataset & Resolution & Categories & Train split & Val Split \\
    \midrule
    Potsdam & $6000\times6000$ & 6 & 24 & 14 \\
    Vaihingen & $\sim 2494\times2064$ & 6 &  16 & 17 \\
    LoveDA & $1024\times1024$ & 7 & 2521 & 1668\\
    iSAID & $800\times800\sim13000\times13000$ &16 & 1403&468 \\
    \bottomrule
    \end{tabular}
    \caption{Dataset overview}
    \label{tab:dataset}
\end{table*}
\begin{table*}[t!]
    \centering
    \setlength{\tabcolsep}{0.8mm} % 设置列间距
    \fontsize{9pt}{9pt}\selectfont
    \begin{tabular}{ccccccccc}
    \toprule
    Spatial Adapter & MoA & Frequency Layer & Cutoff Frequency &  U2R (DA) & V2P (DA) & P2V (DG) & R2U (DG)\\
    \midrule
    \ding{51} & - & - & - & \textbf{48.6} &	60.1 &	61.0 &	58.6  \\
    \ding{51} & \ding{51} & Full  & 0.3 & 47.0 	&\underline{60.9}& 61.3 & 58.1  \\
    \ding{51} & \ding{51} & shallow 3  & 0.3& \underline{48.3} &	60.1 &	\underline{63.8} &	58.0   \\
    \ding{51} & \ding{51} & shallow 6  & 0.3& 47.6 &	60.1 &	\underline{63.8} &	\textbf{59.0}   \\
    \ding{51} & \ding{51} & deep 3  & 0.3& 47.4 &	\textbf{61.6} &	63.7 &	\underline{58.7}   \\
    \ding{51} & \ding{51} & deep 6  & 0.3& 48.0 &	60.6 &	\textbf{64.9} &	58.5   \\
    \midrule 
     \ding{51} & \ding{51} & deep 3 / deep 6 & 0.2 / 0.3 & 50.0 & 62.2 & 64.9 & 59.0\\
        \bottomrule
    \end{tabular}
    \caption{\textbf{Ablation of Frequency Adapter with different activation layers.} 
The results demonstrate that the Earth-Adapter tends to be more beneficial when used in deeper network layers. The results in the gray shadow mean the best performance using deep 3 or 6 layers and 0.2 or 0.3 cutoff frequencies.}
    \label{tab:fft_low}
\end{table*}
\begin{table*}[t!]
    \centering
    \setlength{\tabcolsep}{0.8mm} % 设置列间距
    \fontsize{9pt}{9pt}\selectfont
    \begin{tabular}{ccccccccc}
    \toprule
    Spatial Adapter & MoA & Frequency Layer & Cutoff Frequency &  U2R (DA) & V2P (DA) & P2V (DG) & R2U (DG)\\
    \midrule
    \ding{51} & - & - & - & \textbf{48.6} &	60.1 &	61.0 &	58.6  \\
    \ding{51} & \ding{51} & Full  & 0.3 & 47.0 	&\underline{60.9}& 61.3 & 58.1  \\
    \ding{51} & \ding{51} & shallow 3  & 0.3& \underline{48.3} &	60.1 &	\underline{63.8} &	58.0   \\
    \ding{51} & \ding{51} & shallow 6  & 0.3& 47.6 &	60.1 &	\underline{63.8} &	\textbf{59.0}   \\
    \ding{51} & \ding{51} & deep 3  & 0.3& 47.4 &	\textbf{61.6} &	63.7 &	\underline{58.7}   \\
    \ding{51} & \ding{51} & deep 6  & 0.3& 48.0 &	60.6 &	\textbf{64.9} &	58.5   \\
    \midrule 
     \ding{51} & \ding{51} & deep 3 / deep 6 & 0.2 / 0.3 & 50.0 & 62.2 & 64.9 & 59.0\\
        \bottomrule
    \end{tabular}
    \caption{\textbf{Ablation of Frequency Adapter with different activation layers.} 
The results demonstrate that the Earth-Adapter tends to be more beneficial when used in deeper network layers. The results in the gray shadow mean the best performance using deep 3 or 6 layers and 0.2 or 0.3 cutoff frequencies.}
    \label{tab:fft_low}
\end{table*}
\begin{table}[t!]
    \centering

    \begin{tabular}{cc}
    \toprule
    Lr & 1e-4\\
    Training iters & 20k/40k \\
    Batch size & 4/8\\
    $\alpha$ & 0.99\\
    $\lambda_{da}$ & 0.5\\
    Adapter dim & 16/24/32/64 \\
    Cutoff frequency & 0.2/0.3\\
    Activation layer & pre3/pre6/suf3/suf6\\

        \bottomrule
    \end{tabular}
    \caption{Hyper-parameter overview}
    \label{tab:hyper parameter}
\end{table}
% \documentclass{article}
% \usepackage{booktabs}  % 用于美观表格
% \usepackage{multirow}  % 用于合并行

% \begin{document}

\begin{table}[t!]
    \centering
    \resizebox{1.0\linewidth}{!}{
    \begin{tabular}{ccccc}
        \toprule
        \multirow{2}{*}{Layer}  & \multicolumn{2}{c}{Adapter} & \multicolumn{2}{c}{Earth-Adapter} \\
          & U2R (DA)& R2U (DG)& U2R (DA)& R2U (DG) \\
        \midrule
        Frozen & 11.8 & 57.1 & 11.8 & 57.1\\
        \midrule
       \texttt{[0,1,2,3,4,5]}  & 44.7 & 57.4 & 47.6 & 59.0 \\
       \texttt{[18,19,20,21,22,23]}  & 46.0 & 58.1 & 50.0 & 58.5 \\
        \bottomrule
    \end{tabular}
    }
    \caption{Layer effect comparison between Adapter and Earth-Adapter. The results show Earth-Adapter learns better under the same layer setting.}
    \label{tab:layer_comparision}
\end{table}

% \end{document}

\begin{table}[t]
    \centering
    \resizebox{1.0\linewidth}{!}{
    \begin{tabular}{ccccc}
    \toprule
    Benchmark & Adapter dim & Cutoff frequency & Activation layer & mIoU(\%)\\
    \midrule
        P2V (DA) & 64 & 0.3 & \texttt{[0,1,2]} & 67.7 \\
        P2V (DG)& 24 & 0.3 & \texttt{[18,19,20,21,22,23]} & 64.9 \\
        V2P (DA) & 32 & 0.2 & \texttt{[21,22,23]} & 62.2 \\
        V2P (DG) & 64 & 0.3 & \texttt{[18,19,20,21,22,23]} & 55.1 \\
        R2U (DA) & 16 & 0.3 & \texttt{[21,22,23]} & 55.9 \\
        R2U (DG)& 64 & 0.3 & \texttt{[18,19,20,21,22,23]} & 59.0 \\
        U2R (DA)& 32 & 0.2 & \texttt{[18,19,20,21,22,23]} & 50.0 \\
        U2R (DG)& 64 & 0.3 & \texttt{[21,22,23]} & 45.7 \\
        \bottomrule
    \end{tabular}
    }
    \caption{Hyperparameter Configuration of dimension, cutoff frequency, and activation layer of Frequency Adapters in Earth-Adapter. Notably, Spatial Adapter is activated in all backbone layers. All the results are the best performance.}
    \label{tab:hyper parameter}
\end{table}

We conduct experiments on several major optical remote sensing image segmentation datasets. The original datasets have varying resolutions, ranging from $1024\times1024$ to $6000\times6000$, which we uniformly crop to $512\times512$ for training and evaluation. The detailed information of the datasets are shown in the table below.

\subsection{Training Framework}

In this study, we explore the application of Earth-Adapter across three distinct tasks: In-Domain(SS), DG and DA Semantic Segmentation.
For SS and DG tasks, we train Earth-Adapter with Mask2former in end-to-end supervised learning as Equation~\ref{loss_fun1}.
For DA Semantic Segmentation, we use the self-training framework introduced in DACS~\cite{dacs} without using target domain labels. The pseudo labels for target domain images are produced by a teacher network $G_{\theta, \xi, \epsilon}^\dagger$:
\begin{equation}
    \label{eq:eq3}
    \mathbf{\Bar{y}_T}^{(j,k)} = [k = \mathop{\arg\max}_{k'}G_{\theta, \xi, \epsilon}^\dagger(\mathbf{x_T})))^{(j,k')}],
\end{equation}
where $[\cdot]$ denotes the Iverson bracket, and $j\in HW$ is the spatial index of the pixel in the image.
 The teacher network remains inactive during the training process, with its parameters being updated based on an exponential moving average (EMA) of the student model's parameters: 
\begin{equation}
    G_{\theta, \xi, \epsilon,t+1}^\dagger = \alpha G_{\theta, \xi, \epsilon,t}^\dagger+(1-\alpha)G_{\theta, \xi, \epsilon,t}
\end{equation}
Here $t$ denotes the $t_{th}$ training iteration.
The hyperparameter $\alpha$ controls the update speed of the teacher network's parameters.
After obtaining the pseudo-labels, we perform category-wise mixing of samples from the source and target domains:
\begin{equation}\label{eq:eq14}
\begin{aligned}
     \mathbf{x}_{mix}&= \mathbf{x_S}\odot\mathcal{M} +\mathbf{x_T}\odot(1-\mathcal{M})\\
     \mathbf{y}_{mix}&= \mathbf{y_S}\odot\mathcal{M} +\mathbf{\Bar{y}_T}\odot(1-\mathcal{M})\\
\end{aligned}
\end{equation}
where $(\mathbf{x_S},\mathbf{y_S})$ is the data sample from the source domain, and $\mathcal{M}$ is generated by randomly selecting half of the categories based on the labels of the source domain $\mathbf{y_S}$.
The Optimization objective for the DA task if defined by:
\begin{equation}
    \begin{aligned}
        \mathop{\arg \min}_{\theta, \epsilon, \xi} \, & \sum_{(\mathbf{x}, \mathbf{y}) \in \mathcal{D} \cup \mathcal{D}_{mix}} \mathcal{L}_{\text{seg}}\left(G_{\theta, \xi, \epsilon}(\mathbf{x}), \mathbf{y}\right) \\
        & \quad + \lambda_{uda} \mathcal{L}_{\text{seg}}\left(G_{\theta, \xi, \epsilon}(\mathbf{x}_{mix}), \mathbf{y}_{mix}\right).
    \end{aligned}
\end{equation}

\subsection{Hyper-Parameter Configuration}
As shown in Table 7, we provide an overview of  training hyperparameters. Specifically, for DA and DG tasks, we train for 20k iterations with a batch size of 4; for SS tasks, we train for 40k iterations with a batch size of 8. Notably, for the iSAID SS task, we train for 80k iterations with a batch size of 8. Other configurations are detailed in rows 5 to 7 of the table.
\section{More Ablation and Analysis}

\subsection{Analysis of Overall Parameter Configuration.}
The detailed hyperparameter configurations for each benchmark are presented in Table~\ref{tab:hyper parameter}. Overall, our algorithm demonstrates exceptional robustness, exhibiting relatively low sensitivity to parameter variations. Below we provide an in-depth analysis of the hyperparameter effects.

\textbf{Effect of Adapter dimension.} The Earth-Adapter achieves optimal performance across varying dimensions on different benchmarks: P2V (DA) attains peak performance (67.3\% mIoU at dim=64), P2V (DG) reaches its highest score (64.9\% at dim=24), V2P (DA) achieves 62.2\% mIoU at dim=32, while V2P (DG) attains its best performance at dim=64. A notable pattern emerges where domain generalization (DG) tasks consistently require larger-dimension adapters—V2P (DG) and R2U (DG) both peak at dim=64—whereas domain adaptation (DA) tasks achieve optimal results with smaller dimensions (V2P (DA) at dim=32 and R2U (DA) at dim=16). We attribute this phenomenon to DG tasks demanding greater parameter capacity to  facilitate the foundation model's extraction of more comprehensive semantic representations, thereby significantly boosting cross-domain generalization performance.

\textbf{Effect of cutoff frequency.} The cutoff frequency in the frequency domain serves as the boundary between high and low frequencies. Empirical evidence suggests that a relatively low cutoff value is generally preferable. In our experiments, we evaluated performance at cutoff frequencies of 0.2 and 0.3. Different benchmarks achieved optimal results at different frequencies: six benchmarks (P2V (DA), P2V (DG), V2P (DG), R2U (DA), R2U (DG), and U2R (DG)) attained peak performance at a cutoff ratio of 0.3, while V2P (DA) and U2R (DA) performed best at a ratio of 0.2.

\textbf{Effect of Frequency Adapter layer.} We also conducted experiments on the configuration of frequency-domain adapters across different layers. Our study primarily focused on the Dinov2-Large model, a 24-layer Vision Transformer architecture. We evaluated four distinct adapter configurations: [0,1,2], [0,1,2,3,4,5], [21,22,23], and [18,19,20,21,22,23]. As shown in Table~\ref{tab:hyper parameter}, different benchmarks achieved optimal performance with varying frequency-domain adapter configuration. For instance, P2V (DA) attained its peak performance of 67.7\% mIoU with adapters in [0,1,2], while P2V (DG) achieved its best results with the [18,19,20,21,22,23] configuration. Similar to the adapter dimension findings, the frequency adapter layers demonstrate task-dependent optimization patterns. Notably, three DG tasks (P2V (DG), V2P (DG), and U2R (DG)) consistently performed best with adapters in the layers [18,19,20,21,22,23], suggesting that domain generalization (DG) requires more adapter parameters to capture general semantic representations to improve cross-domain performance.

\textbf{Layer-wise comparison between Adapter and Earth-Adapter.} We analyze the layer-wise behavior of Adapter and compare the performance of Adapter and Earth-Adapter under different layer configurations (Table~\ref{tab:layer_comparision}). While prior work~\cite{dinov2} suggests that deeper layers capture more semantically meaningful features, our Earth-Adapter demonstrates superior performance in both shallow-layer (first six) and deep-layer (last six) fine-tuning. For instance, in the U2R (DA) task, Earth-Adapter (shallow fine-tuning) outperforms the frozen model by 35.8\% mIoU, while in R2U (DG), it surpasses the frozen baseline by 1.9\% mIoU. Similar improvements are observed for deep-layer fine-tuning, achieving +38.2\% mIoU (U2R (DA)) and +1.6\% mIoU (U2R (DG)). Moreover, compared to Adapter, Earth-Adapter consistently shows better adaptation—+2.9\% mIoU (U2R (DA), shallow) and +0.4\% mIoU (R2U (DG), deep). These results highlight Earth-Adapter’s robustness and generalization capability across different layer-wise configurations.

\textbf{Detailed analysis of parameter configuration of Frequency Adapter.} We conduct a comprehensive analysis of the impact of frequency-domain layers on model performance, as shown in Table~\ref{tab:fft_low}. In our default setting, the cutoff frequency of Earth-Adapter is set to 0.3. Our experiments reveal that these frequency layers play a crucial role in the overall performance of Earth-Adapter. Notably, integrating the Frequency Adapter results in a slight performance drop in two benchmarks: U2R (DA) ($-1.6\%$ mIoU) and R2U (DG) ($-0.5\%$ mIoU). However, with an efficient parameter search, Earth-Adapter still surpasses the Spatial-Adapter. As shown in the table, Earth-Adapter achieves 50.0\% mIoU in the U2R (DA) task, outperforming Spatial-Adapter by 1.4\% mIoU. Similarly, it attains 59.0\% mIoU in the R2U (DG) task, exceeding Spatial-Adapter by 0.4\% mIoU. A key observation is that Earth-Adapter exhibits sensitivity to frequency layer configurations and cutoff frequencies under certain benchmarks—which we aim to refine in future work.

\bibliography{aaai2026}

\end{document}